%% file: acl2020.tex
\documentclass[11pt,a4paper]{article}
\usepackage[hyperref]{acl2020}
\usepackage{times}
\usepackage{latexsym}
\usepackage{times}
\usepackage{latexsym}
\usepackage{url}
\usepackage{times}
\usepackage{latexsym}
\usepackage{amsmath}
\usepackage{amsfonts}
\usepackage{xfrac}
\usepackage{mathtools}
\usepackage[table]{colortbl}%
\usepackage{tikz}
\usepackage{caption}
\usepackage{booktabs}
\usepackage{adjustbox}
\usepackage{xspace}
\usepackage[normalem]{ulem}
\usepackage{subfig}
\usepackage{pifont}
\usepackage{bbm}
\usepackage{xcolor}
\usepackage{algorithm}
\usepackage{multirow}
\usepackage[noend]{algpseudocode}

\usepackage{todonotes}

\usepackage{cleveref}
\crefname{section}{\S}{\S\S}
\crefname{table}{Tab.}{}
\crefname{figure}{Fig.}{}
\crefname{algorithm}{Alg.}{}
\crefname{equation}{eq.}{}
\crefname{appendix}{App.}{}
\crefformat{section}{\S#2#1#3}  %

\usepackage{microtype}

\aclfinalcopy %

\input{table}

\title{\raisebox{1ex}[0in][0in]{\parbox[b]{\linewidth}{\begin{flushright}\footnotesize
        \textmd{\textsf{\textcolor{gray}{This paper appeared in the 5$^\text{th}$ RepL4NLP workshop and was presented online in July 2020. This version \\
        was prepared in September 2020 to include breakdown of experiment in \cref{sec:is-mbert-equal}, which can be found in \cref{app:breakdown}.\\
        Code is available at \url{https://github.com/shijie-wu/crosslingual-nlp}.}}}\end{flushright}}}\\ 
        \vspace{-1ex}Are All Languages Created Equal in Multilingual BERT?}

\author{Shijie Wu \and Mark Dredze \\
Department of Computer Science \\
Johns Hopkins University \\
{\tt shijie.wu@jhu.edu, mdredze@cs.jhu.edu}
}

\date{}

\begin{document}
\maketitle
\begin{abstract}
Multilingual BERT (mBERT) \cite{multilingualbertmd} trained on 104 languages has shown surprisingly good cross-lingual performance on several NLP tasks, even without explicit cross-lingual signals \cite{wu-dredze-2019-beto,pires-etal-2019-multilingual}.
However, these evaluations have focused on cross-lingual transfer with high-resource languages, covering only a third of the languages covered by mBERT. We explore how mBERT performs on a much wider set of languages, focusing on the quality of representation for low-resource languages, measured by within-language performance. We consider three tasks: Named Entity Recognition (99 languages), Part-of-speech Tagging, and Dependency Parsing (54 languages each). mBERT does better than or comparable to baselines on high resource languages but does much worse for low resource languages. Furthermore, monolingual BERT models for these languages do even worse. Paired with similar languages, the performance gap between monolingual BERT and mBERT can be narrowed. We find that better models for low resource languages require more efficient pretraining techniques or more data.
\end{abstract}

\section{Introduction}
Pretrained contextual representation models trained with language modeling  \cite{peters-etal-2018-deep,yang2019xlnet} or the cloze task objectives \cite{devlin-etal-2019-bert,liu2019roberta} have quickly set a new standard for NLP tasks.
These models have also been trained in multilingual settings.
As the authors of BERT say ``[...] (they) do not plan to release more single-language models'', they instead train a single BERT model with Wikipedia to serve 104 languages, without any explicit cross-lingual links, yielding a multilingual BERT (mBERT) \cite{multilingualbertmd}. Surprisingly, mBERT learn high-quality cross-lingual representation and show strong zero-shot cross-lingual transfer performance \cite{wu-dredze-2019-beto,pires-etal-2019-multilingual}.
However, evaluations have focused on high resource languages, with cross-lingual transfer using English as a source language or within language performance. As \newcite{wu-dredze-2019-beto} evaluated mBERT on 39 languages, this leaves the majority of mBERT's 104 languages, most of which are low resource languages, untested. 

\textit{Does mBERT learn equally high-quality representation for its 104 languages?} If not, which languages are hurt by its massively multilingual style pretraining? While it has been observed that for high resource languages like English, mBERT performs worse than monolingual BERT on English with the same capacity \cite{multilingualbertmd}. 
It is unclear that for low resource languages (in terms of monolingual corpus size), how does mBERT compare to a monolingual BERT? And, does multilingual joint training help mBERT learn better representation for low resource languages?

We evaluate the representation quality of mBERT on 99 languages for NER, and 54 for part-of-speech tagging and dependency parsing. 
In this paper, we show mBERT does not have equally high-quality representation for all of the 104 languages, with the bottom 30\% languages performing much worse than a non-BERT model on NER. Additionally, by training various monolingual BERT for low-resource languages with the same data size, we show the low representation quality of low-resource languages is not the result of the hyperparameters of BERT or sharing the model with a large number of languages, as monolingual BERT performs worse than mBERT. On the contrary, by pairing low-resource languages with linguistically-related languages, we show low-resource languages benefit from multilingual joint training, as bilingual BERT outperforms monolingual BERT while still lacking behind mBERT.

Our findings suggest, with small monolingual corpus, BERT does not learn high-quality representation for low resource languages. To learn better representation for low resource languages, we suggest either collect more data to make low resource language high resource \cite{conneau2019unsupervised}, or consider more data-efficient pretraining techniques like \newcite{clark2020electra}. We leave exploring more data-efficient pretraining techniques as future work.

\section{Related Work}
\paragraph{Multilingual Contextual Representations}
Deep contextualized representation models such as ELMo \cite{peters-etal-2018-deep} and BERT \cite{devlin-etal-2019-bert} have set a new standard for NLP systems.
Their application to multilingual settings, pretraining one model on text from multiple languages with a single vocabulary, has driven forward work in cross-language learning and transfer \cite{wu-dredze-2019-beto,pires-etal-2019-multilingual,mulcaire-etal-2019-polyglot}. BERT-based pretraining also benefits language generation tasks like machine translation \cite{lample2019cross}.
BERT can be further improve with explicit cross-language signals including: bitext \cite{lample2019cross,huang-etal-2019-unicoder} and word translation pairs from a dictionary \cite{wu2019emerging} or induced from a bitext \cite{ji2019cross}.

Several factors need to be considered in understanding mBERT. First, the 104 most common Wikipedia languages vary considerably in size (Table \ref{tab:lang}). Therefore, mBERT training attempted to equalize languages by up-sampling words from low resource languages and down-sampling words from high resource languages. 
Previous work has found that shared strings across languages provide sufficient signal for inducing cross-lingual word representations \cite{conneau2017word,artetxe-etal-2017-learning}.
While \newcite{wu-dredze-2019-beto} finds the number of shared subwords across languages correlates with cross-lingual performance, multilingual BERT can still learn cross-lingual representation without any vocabulary overlap across languages \cite{wu2019emerging,K2020Cross-Lingual}. Additionally, \newcite{wu2019emerging} find bilingual BERT can still achieve decent cross-lingual transfer by sharing only the transformer layer across languages. \newcite{artetxe2019cross} shows learning the embedding layer alone while using a fixed transformer encoder from English monolingual BERT can also produce decent cross-lingual transfer performance.
Second, while each language may be similarly represented in the training data, subwords are not evenly distributed among the languages. Many languages share common characters and cognates, biasing subword learning to some languages over others. Both of these factors may influence how well mBERT learns representations for low resource languages.

Finally, \citet{baevski-etal-2019-cloze} show that in general 
larger pretraining data for English leads to better downstream performance, yet increasing the size of pretraining data exponentially only increases downstream performance linearly. For a low resource language with limited pretraining data, it is unclear whether contextual representations outperform previous methods.

\paragraph{Representations for Low Resource Languages} Embeddings with subword information, a non-contextual representation, like fastText \cite{bojanowski-etal-2017-enriching} and BPEmb \cite{heinzerling-strube-2018-bpemb} are more data-efficient compared to contextual representation like ELMo and BERT when a limited amount of text is available. For low resource languages, there are usually limits on \textbf{monolingual corpora} and \textbf{task specific supervision}. When task-specific supervision is limited, e.g. sequence labeling in low resource languages, mBERT performs better than fastText while underperforming a single BPEmb trained on all languages \cite{heinzerling-strube-2019-sequence}. Contrary to this work, we focus on mBERT from the perspective of representation learning for each language in terms of monolingual corpora resources and analyze how to improve BERT for low resource languages. We also consider parsing in addition to sequence labeling tasks.

Concurrently, \newcite{conneau2019unsupervised} train a multilingual masked language model \cite{devlin-etal-2019-bert} on 2.5TB of CommonCrawl filtered data covering 100 languages and show it outperforms a Wikipedia-based model on low resource languages (Urdu and Swahili) for XNLI \cite{conneau-etal-2018-xnli}. Using CommonCrawl greatly increases monolingual resource especially for low resource languages, and makes low resource languages in terms of Wikipedia size high resource. For example, Mongolian has 6 million and 248 million tokens in Wikipedia and CommonCrawl, respectively. Indeed, a 40-fold data increase of Mongolian (mn) increases its WikiSize, a measure of monolingual corpus size introduced in \cref{ssec:wikisize}, from 5 to roughly 10, as shown in \cref{tab:lang}, making it relatively high resource with respect to mBERT.

\insertWikiRankTable

\section{Experimental Setup}
We begin by defining high and low resource languages in mBERT, a description of the models and downstream tasks we use for evaluation, followed by a description of the masked language model pretraining.

\subsection{High/Low Resource Languages}\label{ssec:wikisize}
Since mBERT was trained on articles from Wikipedia, a language is considered a high or low resource for mBERT based on the size of Wikipedia in that language. Size can be measured in many ways (articles, tokens, characters); we use the size of the raw dump archive file;\footnote{The size of English (en) is the size of this file: \url{https://dumps.wikimedia.org/enwiki/latest/enwiki-latest-pages-articles.xml.bz2}} for convenience we use $\log_2$ of the size in MB (\textbf{WikiSize}). English is the highest resource language (15.5GB) and Yoruba the lowest (10MB).\footnote{The ordering does not necessarily match the number of speakers for a language.} \cref{tab:lang} shows languages and their relative resources.

\subsection{Downstream Tasks} 
mBERT supports 104 languages, and we seek to evaluate the learned representations for as many of these as possible. We consider three NLP tasks for which annotated task data exists in a large number of languages: named entity recognition (NER), universal part-of-speech (POS) tagging and universal dependency parsing. For each task, we train a task-specific model using within-language supervised data on top of the mBERT representation with fine-tuning.

For NER we use data created by \citet{pan-etal-2017-cross} automatically built from Wikipedia, which covers 99 of the 104 languages supported by mBERT. We evaluate NER with entity-level F1. This data is in-domain as mBERT is pretrained on Wikipedia. For POS tagging and dependency parsing, we use Universal Dependencies (UD) v2.3 \cite{ud2.3}, which covers 54 languages (101 treebanks) supported by mBERT. We evaluate POS with accuracy (ACC) and Parsing with label attachment score (LAS)  and unlabeled attachment score (UAS). For POS, we consider UPOS within the treebank. For parsing, we only consider universal dependency labels. The domain is treebank-specific so we use all treebanks of a language for completeness.

\paragraph{Task Models} For sequence labeling tasks (NER and POS), we add a linear function with a softmax on top of mBERT. For NER, at test time, we adopt a simple post-processing heuristic as a structured decoder to obtain valid named entity spans. Specifically, we rewrite stand-alone prediction of \texttt{I-X} to \texttt{B-X} and inconsistent prediction of \texttt{B-X I-Y} to \texttt{B-Y I-Y}, following the final entity. For dependency parsing, we replace the LSTM in the graph-based parser of \citet{dozat2016deep} with mBERT. For the parser, we use the original hyperparameters. Note we do not use universal part-of-speech tags as input for dependency parsing. We fine-tune all parameters of mBERT for a specific task.
We use a maximum sequence length of 128 for sequence labeling tasks. For sentences longer than 128, we use a sliding window with 64 previous tokens as context. For dependency parsing, we use sequence length 128 due to memory constraints and drop sentences with more than 128 subwords. We also adopt the same treatment for the baseline \cite{che-etal-2018-towards} to obtain comparable results. Since mBERT operates on the subword-level, we select the first subword of each word for the task-specific layer with masking.

\paragraph{Task Optimization} We train all models with Adam \cite{kingma2014adam}. We warm up the learning rate linearly in the first 10\% steps then decrease linearly to 0. We select the hyperparameters based on dev set performance by grid search, as recommended by \citet{devlin-etal-2019-bert}. The search includes a learning rate (2e-5, 3e-5, and 5e-5), batch size (16 and 32). 
As task-specific supervision size differs by language or treebank, we fine-tune the model for 10k gradient steps and evaluate the model every 200 steps.
We select the best model and hyperparameters for a language or treebank by the corresponding dev set.

\paragraph{Task Baselines}
We compare our mBERT models with previously published methods: \citet{pan-etal-2017-cross} for NER; For POS and dependency parsing the best performing system ranked by LAS in the 2018 universal parsing shared task \cite{che-etal-2018-towards}  \footnote{The shared task uses UD v2.2 while we use v2.3. However, treebanks contain minor changes from version to version.}, which use ELMo as well as word embeddings. Additionally, \newcite{che-etal-2018-towards} is trained on POS and dependency parsing jointly while we trained mBERT to perform each task separately. As a result, the dependency parsing with mBERT does not have access to POS tags. By comparing mBERT to these baselines, we control for task and language-specific supervised training set size.

\subsection{Masked Language Model Pretraining}
We include several experiments in which we pretrain BERT from scratch. We use the PyTorch \cite{paszke2019pytorch} implementation by \newcite{lample2019cross}.\footnote{\url{https://github.com/facebookresearch/XLM}} All sentences in the corpus are concatenated. For each language, we sample a batch of $N$ sequence and each sequence contains $M$ tokens, ignoring sentence boundaries. When considering two languages, we sample each language uniformly. We then randomly select 15\% of the input tokens for masking, proportionally to the exponentiated token count of power -0.5, favoring rare tokens. We replace selected masked token with \texttt{<MASK>} 80\% of the time, the original token 10\% of the time, and uniform random token within the vocabulary 10\% of the time. The model is trained to recover the original token \cite{devlin-etal-2019-bert}. We drop the next sentence prediction task as \newcite{liu2019roberta} find it does not improve downstream performance.

\paragraph{Data Processing}
We extract text from a Wikipedia dump with Gensim \cite{rehurek_lrec}. We learn vocabulary for the corpus using SentencePiece \cite{kudo-richardson-2018-sentencepiece} with the unigram language model \cite{kudo-2018-subword}. When considering two languages, we concatenate the corpora for the two languages while sampling the same number of sentences from both corpora when learning vocabulary. We learn a vocabulary of size $V$, excluding special tokens. Finally, we tokenized the corpora using the learned SentencePiece model and did not apply any further preprocessing.

\paragraph{BERT Models} Following mBERT, We use 12 Transformer layers \cite{vaswani2017attention} with 12 heads, embedding dimensions of 768, hidden dimension of the feed-forward layer of 3072, dropout of 0.1 and GELU activation \cite{hendrycks2016bridging}. We tied the output softmax layer and input embeddings \cite{press-wolf-2017-using}. We consider both a 12 layer model (\textbf{base}) and a smaller 6 layer model (\textbf{small}).

\paragraph{BERT Optimization} We train BERT with Adam and an inverse square root learning rate scheduler with warmup \cite{vaswani2017attention}. We warm up linearly for 10k steps and the learning rate is 0.0001. We use batch size $N=88$ and mixed-precision training. We trained the model for roughly 115k steps and save a checkpoint every 23k steps, which correspond to 10 epochs. We select the best out of five checkpoints with a task-specific dev set. We train each model on a single NVIDIA RTX Titan with 24GB of memory for roughly 20 hours.

\section{Are All Languages Created Equal in mBERT?}\label{sec:is-mbert-equal}

\begin{figure*}[t]
\centering
\includegraphics[width=2\columnwidth]{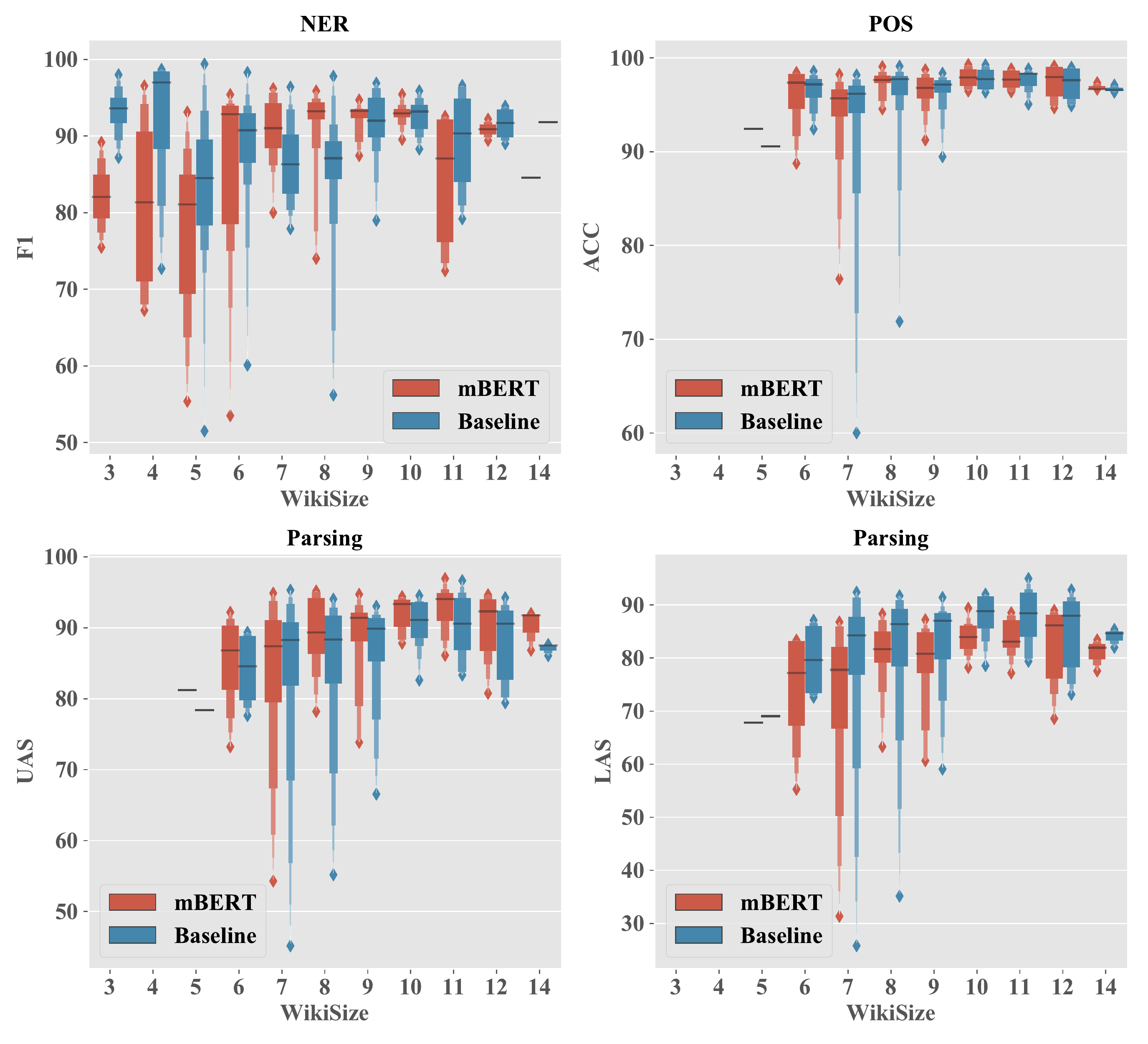}
\caption{mBERT vs baseline grouped by WikiSize. mBERT performance drops much more than baseline models on languages lower than WikiSize 6 -- the bottom 30\% languages supported by mBERT -- especially in NER, which covers nearly all mBERT supported languages. Breakdown can be found in \cref{app:breakdown}.}
\label{fig:baseline}
\end{figure*}

\begin{figure}[t]
\centering
\includegraphics[width=1\columnwidth]{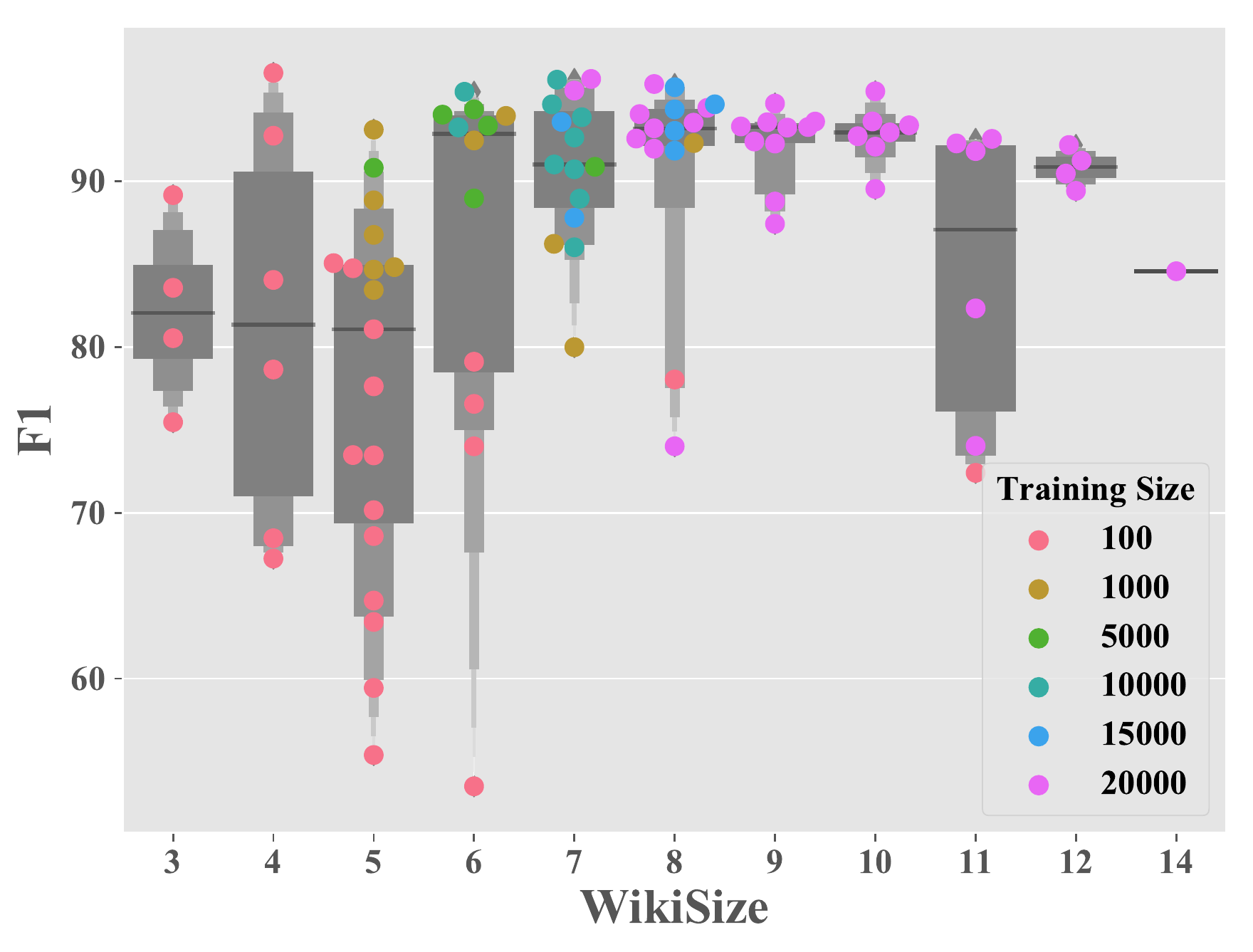}
\caption{NER with mBERT on 99 languages, ordered by size of pretraining corpus (WikiSize). Task-specific supervised training size differs by language. Performance drops dramatically with less pretraining and supervised training data.}
\label{fig:ner-only}
\end{figure}

\cref{fig:baseline} shows the performance of mBERT and the baseline averaged across all languages by Wikipedia size (see \cref{tab:lang} for groupings). 
For WikiSize over 6, mBERT is comparable or better than baselines in all three tasks, with the exception of NER. For NER in very high resource languages (WikiSize over 11, i.e. top 10\%) mBERT performs worse than baseline, suggesting high resource languages could benefit from monolingual pretraining. Note mBERT has strong UAS on parsing but weak LAS compared to the baseline; \citet{wu-dredze-2019-beto} finds adding POS to mBERT improve LAS significantly. We expect multitask learning on POS and parsing could further improve LAS.
While POS and Parsing only cover half (54) of the languages, NER covers 99 of 104 languages, extending the curve to the lowest resource languages. mBERT performance drops significantly for languages with WikiSize less than 6 (bottom 30\% languages). For the smallest size, mBERT goes from being competitive with state-of-the-art to being {\em over 10 points behind.} Readers may find this surprising since while these are very low resource languages, mBERT training up-weighted these languages to counter this effect.

\cref{fig:ner-only} shows the performance of mBERT (only) for NER over languages with {\em different resources}, where we show how much task-specific supervised training data was available for each language. For languages with only 100 labeled sentences, the performance of mBERT drops significantly as these languages also had less pretraining data.
While we may expect that pretraining representations with mBERT would be most beneficial for languages with only 100 labels, as \newcite{howard-ruder-2018-universal} show pretraining improve data-efficiency for English on text classification, our results show that on low resource languages this strategy performs much worse than a model trained directly on the available task data.
Clearly, mBERT provides variable quality representations depending on the language. While we confirm the finding of others that mBERT is excellent for high resource languages, it is much worse for low resource languages.
Our results suggest caution for those expecting a reliable model for \textit{all} 104 mBERT languages.

\section{Why Are All Languages Not Created Equal in mBERT?}\label{sec:why-not-equal}

\subsection{Statistical Analysis}\label{sec:stat-test}

We present a statistical analysis to understand why mBERT does so poorly on some languages.
We consider three factors that might affect the downstream task performance: pretraining Wikipedia size (WikiSize), task-specific supervision size, and vocabulary size in task-specific data. Note we take $\log_2$ of training size and training vocab following WikiSize.
We consider NER because it covers nearly all languages of mBERT.

\insertSigTestTable

We fit a linear model to predict task performance (F1) using a single factor. \cref{tab:sigtest} shows that each factor has a statistically significant positive correlation. One unit increase of training size leads to the biggest performance increase, then training vocabulary followed by WikiSize, all in log scale. 
Intuitively, training size and training vocab correlate with each other. We confirm this with a log-likelihood ratio test; adding training vocabulary to a linear model with training size yields a statistically insignificant improvement. As a result, when considering multiple factors, we consider training size and WikiSize. Interestingly, \cref{tab:sigtest} shows training size still has a positive but slightly smaller slope, but the
slope of WikiSize change sign, which suggests WikiSize might correlate with training size. We confirm this by fitting a linear model with training size as $x$ and WikiSize as $y$ and the slope is over 0.5 with
$p<0.001$. This finding is unsurprising as the NER dataset is built from Wikipedia so larger Wikipedia size means larger training size.

In conclusion, the larger the task-specific supervised dataset, the better the downstream performance on NER. Unsurprisingly, while pretraining improve data-efficiency \cite{howard-ruder-2018-universal}, it still cannot solve a task with limited supervision.
Training vocabulary and Wikipedia size correlate with training size, and increasing either one factor leads to better performance. A similar conclusion could be found when we try to predict the performance ratio of mBERT and the baseline instead.
Statistical analysis shows a correlation between resource and mBERT performance but can not give a causal answer on why low resource languages within mBERT perform poorly.

\subsection{mBERT vs monolingual BERT}
\label{sec:monolingual-bert}

We have established that mBERT does not perform well in low-resource languages. Is this because we are relying on a multilingual model that favors high-resource over low-resource languages? To answer this question we train mono-lingual BERT models on several low resource languages with different hyperparameters. Since pretraining a BERT model from scratch is computationally intensive, we select four low resource languages: Latvian (lv), Afrikaans (af), Mongolian (mn), and Yoruba (yo). These four languages (bold font in \cref{tab:langstat}) reflect varying amounts of monolingual training data. 

\insertLangStatTable

\begin{figure*}[t]
\centering
\includegraphics[width=2\columnwidth]{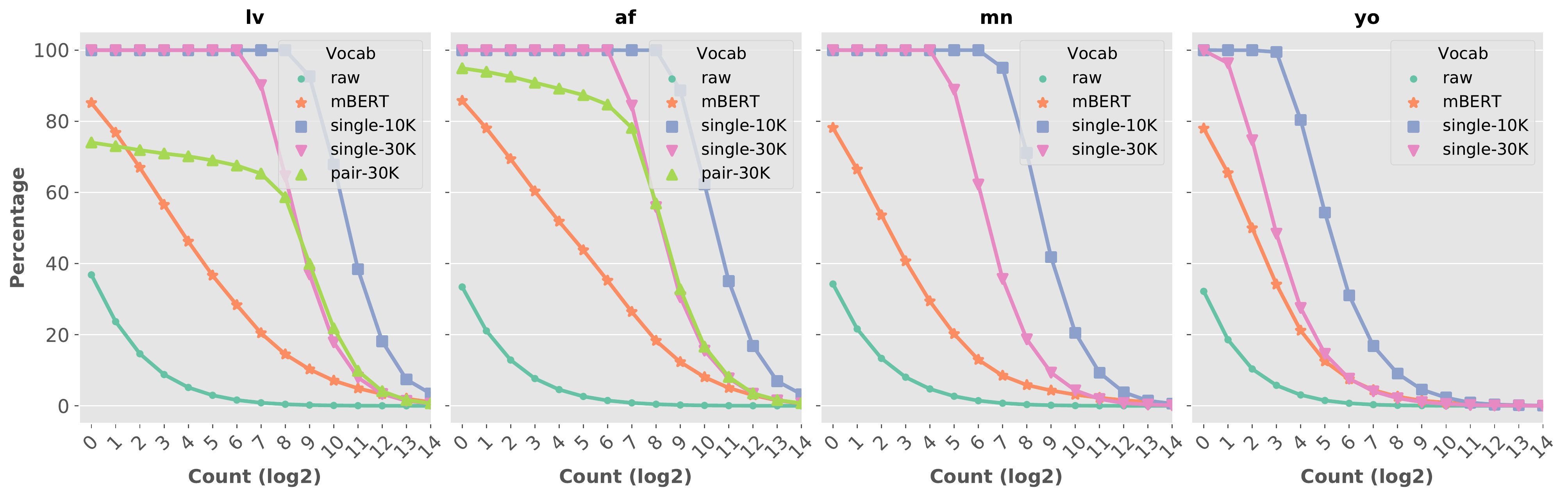}
\caption{Percentage of vocabulary containing word count larger than a threshold. ``Raw'' is the vocabulary segmented by space. Single-30K and Single-10K are 30K/10K vocabularies learned from single languages. Pair-30K is 30K vocabulary learned from the selected language and a closely related language, described in \cref{sec:bilingual-bert}.}
\label{fig:vocab}
\end{figure*}

It turns out that these low resource languages are reasonably covered by mBERT's vocabulary: 25\% to 50\% of the subword types within the mBERT 115K vocabulary appear in these languages' Wikipedia. However, the mBERT vocabulary is by no means optimal for these languages. \cref{fig:vocab} shows that a large amount of the mBERT vocabulary that appears in these languages is low frequency while the language-specific SentencePiece vocabulary has a much higher frequency. In other words, the vocabulary of mBERT is not distributed uniformly.

\insertAllTable

\begin{figure*}[t]
\centering
\includegraphics[width=2\columnwidth]{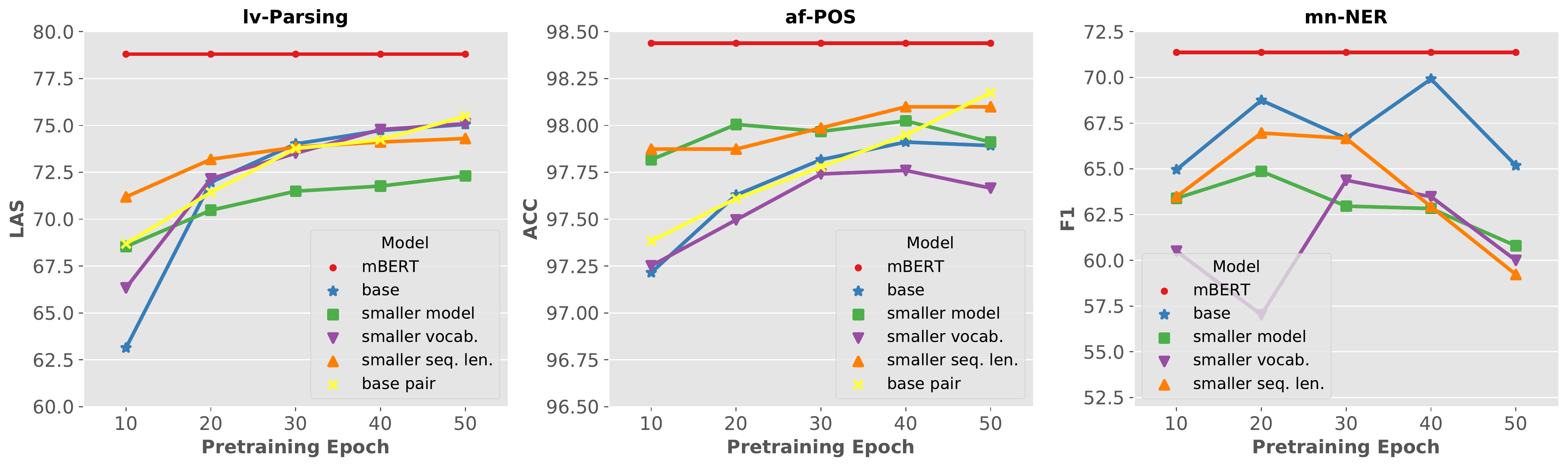}
\caption{Dev performance with different pretraining epochs on three languages and tasks. Dev performance on higher resources languages (lv, af) improves as training continues, while lower resource languages (mn) fluctuate.}
\label{fig:epoch}
\end{figure*}

To train the monolingual BERTs properly for low resource languages, we consider four different sets of hyperparameters. In \textbf{base}, we follow English monolingual BERT on learning vocabulary size $V=30K$, 12 layers of transformer (base). To ensure we have a reasonable batch size for training using our GPU, we set the training sequence length to $M=256$.
Since a smaller model can prevent overfitting smaller datasets, we consider 6 transformer layers (\textbf{small}). We do not change the batch size as a larger batch is observed to improve performance \cite{liu2019roberta}.
As low resource languages have small corpora, 30K vocabulary items might not be optimal. We consider \textbf{smaller vocabulary} with $V=10K$.
Finally, since in fine-tuning we only use a maximum sequence length of 128, in \textbf{smaller sequence length}, we match the fine-tuning phrase with $M=128$. As a benefit of half the self-attention range, we can increase the batch size over 2.5 times to $N=220$.

\cref{tab:all} shows the performance of monolingual BERT in four settings.
The model with smaller sequence length performs best for monolingual BERT and outperforms the base model in 5 out of 8 tasks and languages combination.
The model with smaller vocabulary has mixed performance in the low resource languages (mn, yo) but falls short for (relatively) higher resource languages (lv, af).
Finally, the smaller model underperforms the base model in 5 out of 8 cases.
In conclusion, the best way to pretrain BERT with a limited amount of computation for low resource languages is to use a smaller sequence length to allow a larger batch size. Future work could look into a smaller self-attention span with a restricted transformer \cite{vaswani2017attention} to improve training efficiency.

Despite these insights, no monolingual BERT outperforms mBERT (except Latvian POS). For higher resource languages (lv, af) we hypothesize that training longer with larger batch size could further improve the downstream performance as the cloze task dev perplexity was still improving. \cref{fig:epoch} supports this hypothesis showing downstream dev performance of lv and af improves as pretraining continues. Yet for lower resource languages (mn, yo), the cloze task dev perplexity is stuck and we began to overfit the training set. At the same time, \cref{fig:epoch} shows the downstream performance of mn fluctuates. It suggests the cloze task dev perplexity correlates with downstream performance when dev perplexity is not decreasing.

The fact that monolingual BERT underperforms mBERT on four low resource languages suggests that mBERT style multilingual training benefits low resource languages by transferring from other languages; monolingual training produces worse representations due to small corpus size. Additionally, the poor performance of mBERT on low resource languages does not emerge from balancing between languages. Instead, it appears that we do not have sufficient data, or the model is not sufficiently data-efficient.

\subsection{mBERT vs Bilingual BERT} \label{sec:bilingual-bert}
Finally, we consider a middle ground between monolingual training and massively multilingual training. We train a BERT model on a low resource language (lv and af) paired with a related higher resource language. We pair Lithuanian (lt) with Latvian and Dutch (nl) with Afrikaans.\footnote{We did not consider mn and yo since neither has a closely related language in mBERT.} Lithuanian has a similar size to Latvian while Dutch is over 10 times bigger. Lithuanian belong to the same Genus as Latvian while Afrikaans is a daughter language of Dutch. The \textbf{base pair} model has the same hyperparameters as the base model.

\cref{tab:all} shows that pairing low resource languages with closely related languages improves downstream performance. The Afrikaans-Dutch BERT improves more compared to Latvian-Lithuanian, possibly because Dutch is much larger than Afrikaans, as compared to Latvian and Lithuanian. These experiments suggest that pairing linguistically related languages can benefit representation learning and adding extra languages can further improve the performance as demonstrated by mBERT. It echos the finding of \newcite{lample2019cross} where multilingual training improves uni-directional language model perplexity for low resource languages. Concurrent work shows similar findings as the performance of low resource languages (Urdu and Swahili) improves on XNLI when more languages are trained jointly then decrease with an increasing number of languages \cite{conneau2019unsupervised}. However, they do not consider the effect of language similarity.

\section{Discussion}
While mBERT covers 104 languages, the 30\% languages with least pretraining resources perform worse than using no pretrained language model at all. Therefore, we caution against using mBERT alone for low resource languages. 
Furthermore, training a monolingual model on low resource languages does no better. Training on pairs of closely related low resource languages helps but still lags behind mBERT.
On the other end of the spectrum, the highest resource languages (top 10\%) are hurt by massively multilingual joint training. While mBERT has access to numerous languages, the resulting model is worse than a monolingual model when sufficient training data exists.

Developing pretrained language models for low-resource languages remains an open challenge.
Future work should consider more efficient pretraining techniques, how to obtain more data for low resource languages,
and how to best make use of multilingual corpora.

\section*{Acknowledgments}
This research is supported in part by ODNI, IARPA, via the BETTER Program contract \#2019-19051600005. The views and conclusions contained herein are those of the authors and should not be interpreted as necessarily representing the official policies, either expressed or implied, of ODNI, IARPA, or the U.S. Government. The U.S. Government is authorized to reproduce and distribute reprints for governmental purposes notwithstanding any copyright annotation therein.

\bibliography{anthology,acl2020}
\bibliographystyle{acl_natbib}

\appendix

\section{Breakdown of experiments in \cref{sec:is-mbert-equal}} \label{app:breakdown}

Breakdown of NER in \cref{sec:is-mbert-equal} can be found in \cref{tab:breakdown-ner}, and breakdown of POS tagging and parsing in \cref{sec:is-mbert-equal} can be found in \cref{tab:breakdown-ud}.

\insertNERBreakdownTable

\insertUDBreakdownTable

\end{document}

%% file: table.tex
\newcommand{\insertWikiRankTable}{
\begin{table*}[th]
\begin{center}
\resizebox{1\linewidth}{!}{
\begin{tabular}[b]{cccc}
\toprule
WikiSize & Languages & \# Languages & Size Range (GB) \\
\midrule
3 & io, pms, scn, \textbf{yo} & 4 & [0.006, 0.011] \\
4 & cv, lmo, mg, min, su, vo & 6 & [0.011, 0.022] \\
5 & an, bar, br, ce, fy, ga, gu, is, jv, ky, lb, \textbf{mn}, my, nds, ne, pa, pnb, sw, tg & 19 & [0.022, 0.044] \\
6 & \textbf{af}, ba, cy, kn, la, mr, oc, sco, sq, tl, tt, uz & 12 & [0.044, 0.088] \\
7 & az, bn, bs, eu, hi, ka, kk, lt, \textbf{lv}, mk, ml, nn, ta, te, ur & 15 & [0.088, 0.177] \\
8 & ast, be, bg, da, el, et, gl, hr, hy, ms, sh, sk, sl, th, war & 15 & [0.177, 0.354] \\
9 & fa, fi, he, id, ko, no, ro, sr, tr, vi & 10 & [0.354, 0.707] \\
10 & ar, ca, cs, hu, nl, sv, uk & 7 & [0.707, 1.414] \\
11 & ceb, it, ja, pl, pt, zh & 6 & [1.414, 2.828] \\
12 & de, es, fr, ru & 4 & [2.828, 5.657] \\
14 & en & 1 & [11.314, 22.627] \\
\bottomrule
\end{tabular}
}
\caption{List of 99 languages we consider in mBERT and its pretraining corpus size\label{tab:lang}. Languages in \textbf{bold} are the languages we consider in \cref{sec:why-not-equal}.}
\end{center}
\end{table*}
}

\newcommand{\insertSigTestTable}{
\begin{table}[t]
\begin{center}
\resizebox{1\linewidth}{!}{
\begin{tabular}[b]{cccc}
\toprule
& Coefficient &	p-value & CI \\
\midrule
\multicolumn{4}{l}{\textit{Univariate}} \\
\midrule
Training Size & 0.035 & $<$0.001 & [0.029, 0.041] \\
Training Vocab & 0.021 & $<$0.001 & [0.017, 0.025] \\
WikiSize & 0.015 & $<$0.001 & [0.007, 0.023] \\
\midrule 
\multicolumn{4}{l}{\textit{Multivariate}} \\
\midrule
Training Size & 0.029 & $<$0.001 & [0.023, 0.035] \\
WikiSize & -0.014 & $<$0.001 & [-0.022, -0.006] \\
\bottomrule
\end{tabular}
}
\caption{Statistical analysis on what factors predict downstream performance. We fit two types of linear models, which consider either single factor or multiple factors.
\label{tab:sigtest}}
\end{center}
\end{table}
}

\newcommand{\insertLangStatTable}{
\begin{table}[t]
\begin{center}
\resizebox{1\linewidth}{!}{
\begin{tabular}[b]{c cccc}
\toprule
& lv & af & mn & yo \\
\midrule
Genus & Baltic & Germanic & Mongolic & Defoid \\
Family & Indo-Eur  & Indo-Eur  & Altaic & Niger-Congo \\
WikiSize & 7 & 6 & 5 & 3 \\
\# Sentences (M) & 2.9 & 2.3 & 0.8 & 0.1 \\
\# Tokens (M) & 21.8 & 28.8 & 6.4 & 0.9 \\
mBERT vocab (K) & 56.6 & 59.0 & 42.3 & 29.3 \\
mBERT vocab (\%) & 49.2 & 51.3 & 36.8 & 25.5 \\
\bottomrule
\end{tabular}
}
\caption{Statistic of four low resource languages.
\label{tab:langstat}}
\end{center}
\end{table}
}

\newcommand{\insertAllTable}{
\begin{table*}[t]
\begin{center}
\resizebox{1\linewidth}{!}{
\begin{tabular}[b]{ccc| ccc| ccc| c| c}
\toprule
\multirow{2}{*}{Model Size} & \multirow{2}{*}{Vocabulary} & \multirow{2}{*}{Max Length} & \multicolumn{3}{c|}{lv} & \multicolumn{3}{c|}{af} & \multicolumn{1}{c|}{mn} & \multicolumn{1}{c}{yo} \\
& & & NER & POS & Parsing (LAS/UAS) & NER & POS & Parsing (LAS/UAS) & NER & NER \\
\midrule 
\multicolumn{11}{l}{\textit{Baseline}} \\
\midrule
\multicolumn{3}{c|}{Baseline} & 92.10 & \textbf{96.19} & \textbf{84.47}/88.28 & \textbf{94.00} & 97.50 & \textbf{85.69}/88.67 & \textbf{76.40} & \textbf{94.00} \\
\multicolumn{3}{c|}{mBERT} & \textbf{93.88} & 95.69 & 77.78/\textbf{88.69} & 93.36 & \textbf{98.26} & 83.18/\textbf{89.69} & 64.71 & 80.54 \\
\midrule 
\multicolumn{11}{l}{\textit{Monolingual BERT} (\cref{sec:monolingual-bert})} \\
\midrule
base & 30k & 256  & 93.02 & \underline{95.76} & \underline{74.18}/\underline{85.35} & 90.90 & 97.76 & 80.08/86.92 & 56.20 & 72.57 \\
\midrule
small & - & - & 92.75 & 95.41 & 71.67/83.34 & 90.67 & \underline{98.02} & 80.60/87.40 & \underline{58.92} & 70.80 \\
- & 10k & - & 92.68 & 95.65 & 73.94/85.20 & 89.55 & 97.66 & 79.91/86.93 & 41.70 & \underline{80.18} \\
- & - & 128 & \underline{93.38} & 95.57 & 73.21/84.53 & \underline{91.84} & 97.87 & \underline{80.83}/\underline{87.59} & 55.91 & 73.45 \\
\midrule 
\multicolumn{3}{l}{\textit{Bilingual BERT} (\cref{sec:bilingual-bert})} & \multicolumn{3}{c}{lv + lt} & \multicolumn{3}{c}{af + nl} & \multicolumn{2}{c}{} \\
\midrule
base & 30k & 256  & 93.22 & 96.03 & 74.42/85.60 & 91.85 & 97.98 & 81.73/88.55 & n/a & n/a  \\
\bottomrule
\end{tabular}
}
\caption{Monolingual BERT on four languages with different hyperparameters. \underline{Underscore} denotes best within monolingual BERT and \textbf{bold} denotes best among all models. Monolingual BERT underperforms mBERT in most cases. ``-'' denotes same as base case. \label{tab:all}}
\end{center}
\end{table*}
}

\newcommand{\insertNERBreakdownTable}{
\begin{table*}[t]
\begin{center}
\resizebox{0.75\linewidth}{!}{
\begin{tabular}[b]{cccc|cccc}
\toprule
\textbf{Language} & \textbf{WikiSize} & \textbf{Baseline} & \textbf{mBERT} & \textbf{Language} & \textbf{WikiSize} & \textbf{Baseline} & \textbf{mBERT} \\
\midrule 
io & 3 & 87.2 & \bf 89.2 & ast & 8 & 89.2 & \bf 92.3 \\
pms & 3 & \bf 98.0 & 83.6 & be & 8 & 84.1 & \bf 91.9 \\
scn & 3 & \bf 93.2 & 75.5 & bg & 8 & 65.8 & \bf 93.5 \\
yo & 3 & \bf 94.0 & 80.5 & da & 8 & 87.1 & \bf 93.2 \\
\cellcolor{lightgray} \textbf{AVERAGE} & \cellcolor{lightgray} & \cellcolor{lightgray} \bf 93.1 & \cellcolor{lightgray} 82.2 & el & 8 & 84.6 & \bf 92.0 \\
cv & 4 & \bf 95.7 & 78.7 & et & 8 & 86.8 & \bf 93.1 \\
lmo & 4 & \bf 98.3 & 92.8 & gl & 8 & 87.4 & \bf 94.7 \\
mg & 4 & \bf 98.7 & 96.6 & hr & 8 & 82.8 & \bf 92.6 \\
min & 4 & \bf 85.8 & 68.5 & hy & 8 & 90.4 & \bf 95.7 \\
su & 4 & \bf 72.7 & 67.2 & ms & 8 & 86.8 & \bf 94.4 \\
vo & 4 & \bf 98.5 & 84.1 & sh & 8 & \bf 97.8 & 95.9 \\
\cellcolor{lightgray} \textbf{AVERAGE} & \cellcolor{lightgray} & \cellcolor{lightgray} \bf 91.6 & \cellcolor{lightgray} 81.3 & sk & 8 & 87.3 & \bf 94.1 \\
an & 5 & 93.0 & \bf 93.1 & sl & 8 & 89.5 & \bf 94.3 \\
bar & 5 & \bf 97.1 & 84.8 & th & 8 & 56.2 & \bf 74.0 \\
br & 5 & \bf 87.0 & 84.8 & war & 8 & \bf 94.9 & 78.0 \\
ce & 5 & \bf 99.4 & 85.1 & \cellcolor{lightgray} \textbf{AVERAGE} & \cellcolor{lightgray} & \cellcolor{lightgray} 84.7 & \cellcolor{lightgray} \bf 91.3 \\
fy & 5 & 86.6 & \bf 86.8 & fa & 9 & \bf 96.4 & 93.6 \\
ga & 5 & \bf 85.3 & 83.5 & fi & 9 & \bf 93.4 & 92.4 \\
gu & 5 & \bf 76.0 & 55.4 & he & 9 & 79.0 & \bf 87.4 \\
is & 5 & 80.2 & \bf 84.7 & id & 9 & 87.8 & \bf 93.6 \\
jv & 5 & \bf 82.6 & 73.5 & ko & 9 & \bf 90.6 & 88.8 \\
ky & 5 & 71.8 & \bf 73.5 & no & 9 & \bf 94.1 & 93.3 \\
lb & 5 & 81.5 & \bf 90.8 & ro & 9 & 90.6 & \bf 94.7 \\
mn & 5 & \bf 76.4 & 64.7 & sr & 9 & \bf 95.3 & 93.3 \\
my & 5 & 51.5 & \bf 70.2 & tr & 9 & \bf 96.9 & 93.3 \\
nds & 5 & \bf 84.5 & 81.1 & vi & 9 & 89.6 & \bf 92.3 \\
ne & 5 & \bf 81.5 & 77.6 & \cellcolor{lightgray} \textbf{AVERAGE} & \cellcolor{lightgray} & \cellcolor{lightgray} 91.4 & \cellcolor{lightgray} \bf 92.3 \\
pa & 5 & \bf 74.8 & 59.4 & ar & 10 & 88.3 & \bf 89.5 \\
pnb & 5 & \bf 90.8 & 63.4 & ca & 10 & 90.3 & \bf 93.6 \\
sw & 5 & \bf 93.4 & 88.9 & cs & 10 & \bf 94.6 & 92.7 \\
tg & 5 & \bf 88.3 & 68.6 & hu & 10 & \bf 95.9 & 93.4 \\
\cellcolor{lightgray} \textbf{AVERAGE} & \cellcolor{lightgray} & \cellcolor{lightgray} \bf 83.2 & \cellcolor{lightgray} 77.4 & nl & 10 & \bf 93.2 & 92.1 \\
af & 6 & 85.7 & \bf 93.4 & sv & 10 & 93.6 & \bf 95.4 \\
ba & 6 & \bf 93.8 & 74.0 & uk & 10 & 91.5 & \bf 93.0 \\
cy & 6 & 90.7 & \bf 93.3 & \cellcolor{lightgray} \textbf{AVERAGE} & \cellcolor{lightgray} & \cellcolor{lightgray} 92.5 & \cellcolor{lightgray} \bf 92.8 \\
kn & 6 & \bf 60.1 & 53.5 & ceb & 11 & \bf 96.3 & 72.4 \\
la & 6 & 90.8 & \bf 94.0 & it & 11 & \bf 96.6 & 92.3 \\
mr & 6 & 82.4 & \bf 89.0 & ja & 11 & \bf 79.2 & 74.0 \\
oc & 6 & \bf 92.5 & 76.6 & pl & 11 & 90.0 & \bf 91.8 \\
sco & 6 & \bf 86.8 & 79.1 & pt & 11 & 90.7 & \bf 92.6 \\
sq & 6 & 94.1 & \bf 94.4 & zh & 11 & 82.0 & \bf 82.3 \\
tl & 6 & 92.7 & \bf 95.4 & \cellcolor{lightgray} \textbf{AVERAGE} & \cellcolor{lightgray} & \cellcolor{lightgray} \bf 89.1 & \cellcolor{lightgray} 84.3 \\
tt & 6 & 87.7 & \bf 92.5 & de & 12 & 89.0 & \bf 90.5 \\
uz & 6 & \bf 98.3 & 94.0 & es & 12 & \bf 93.9 & 92.2 \\
\cellcolor{lightgray} \textbf{AVERAGE} & \cellcolor{lightgray} & \cellcolor{lightgray} \bf 88.0 & \cellcolor{lightgray} 85.8 & fr & 12 & \bf 93.3 & 91.3 \\
az & 7 & 85.1 & \bf 91.0 & ru & 12 & \bf 90.1 & 89.4 \\
bn & 7 & 93.8 & \bf 96.1 & \cellcolor{lightgray} \textbf{AVERAGE} & \cellcolor{lightgray} & \cellcolor{lightgray} \bf 91.6 & \cellcolor{lightgray} 90.8 \\
bs & 7 & 84.8 & \bf 93.6 & en & 14 & \bf 91.8 & 84.6 \\
eu & 7 & 82.5 & \bf 92.7 & \cellcolor{lightgray} \textbf{AVERAGE} & \cellcolor{lightgray} & \cellcolor{lightgray} \bf 91.8 & \cellcolor{lightgray} 84.6 \\
hi & 7 & 86.9 & \bf 90.9 &  &  &  &  \\
ka & 7 & 79.8 & \bf 89.0 &  &  &  &  \\
kk & 7 & \bf 88.3 & 86.2 &  &  &  &  \\
lt & 7 & 86.3 & \bf 90.7 &  &  &  &  \\
lv & 7 & 92.1 & \bf 93.9 &  &  &  &  \\
mk & 7 & 93.4 & \bf 94.7 &  &  &  &  \\
ml & 7 & 82.4 & \bf 86.0 &  &  &  &  \\
nn & 7 & 88.1 & \bf 95.5 &  &  &  &  \\
ta & 7 & 77.9 & \bf 87.8 &  &  &  &  \\
te & 7 & \bf 80.5 & 80.0 &  &  &  &  \\
ur & 7 & \bf 96.4 & 96.2 &  &  &  &  \\
\cellcolor{lightgray} \textbf{AVERAGE} & \cellcolor{lightgray} & \cellcolor{lightgray} 86.6 & \cellcolor{lightgray} \bf 91.0 &  &  &  &  \\
\bottomrule
\end{tabular}
}
\caption{Breakdown of NER experiment in \cref{sec:is-mbert-equal}. \label{tab:breakdown-ner}}
\end{center}
\end{table*}
}

\newcommand{\insertUDBreakdownTable}{
\begin{table*}[t]
\begin{center}
\resizebox{0.7\linewidth}{!}{
\begin{tabular}[b]{cc|cc|cc|cc}
\toprule
 &  & \multicolumn{2}{c|}{\textbf{POS (ACC)}} & \multicolumn{2}{c|}{\textbf{Parsing (UAS)}} & \multicolumn{2}{c}{\textbf{Parsing (LAS)}} \\
\textbf{Language \& Treebank} & \textbf{WikiSize} & \textbf{Baseline} & \textbf{mBERT} & \textbf{Baseline} & \textbf{mBERT} & \textbf{Baseline} & \textbf{mBERT} \\
\midrule 

Irish-IDT & 5 & 90.6 & \bf 92.4 & 78.4 & \bf 81.2 & \bf 69.1 & 67.8 \\
\rowcolor{lightgray} \textbf{AVERAGE} & & 90.6 & \bf 92.4 & 78.4 & \bf 81.2 & \bf 69.1 & 67.8 \\

Afrikaans-AfriBooms & 6 & 97.5 & \bf 98.3 & 88.7 & \bf 89.7 & \bf 85.7 & 83.2 \\
Latin-ITTB & 6 & \bf 98.6 & 98.4 & 89.4 & \bf 92.2 & \bf 87.1 & 83.4 \\
Latin-PROIEL & 6 & \bf 96.9 & 96.5 & 77.6 & \bf 83.9 & \bf 73.6 & 71.2 \\
Latin-Perseus & 6 & \bf 92.4 & 88.8 & \bf 80.5 & 73.2 & \bf 72.6 & 55.3 \\
\rowcolor{lightgray} \textbf{AVERAGE} & & \bf 96.3 & 95.5 & 84.0 & \bf 84.8 & \bf 79.8 & 73.3 \\

Basque-BDT & 7 & \bf 96.5 & 96.1 & \bf 87.5 & 87.2 & \bf 84.2 & 76.9 \\
Hindi-HDTB & 7 & \bf 97.6 & 97.2 & \bf 95.3 & 94.9 & \bf 92.4 & 86.8 \\
Kazakh-KTB & 7 & 60.0 & \bf 76.4 & 45.2 & \bf 54.3 & 25.8 & \bf 31.3 \\
Latvian-LVTB & 7 & \bf 96.2 & 95.7 & 88.3 & \bf 88.7 & \bf 84.5 & 77.8 \\
Norwegian-Nynorsk & 7 & 98.2 & \bf 98.2 & 92.7 & \bf 93.5 & \bf 91.0 & 85.7 \\
Norwegian-NynorskLIA & 7 & \bf 94.0 & 93.4 & \bf 76.2 & 71.7 & \bf 70.4 & 56.5 \\
Urdu-UDTB & 7 & \bf 94.2 & 94.0 & \bf 88.9 & 87.4 & \bf 83.4 & 78.5 \\
\rowcolor{lightgray} \textbf{AVERAGE} & & 91.0 & \bf 93.0 & 82.0 & \bf 82.5 & \bf 76.0 & 70.5 \\

Armenian-ArmTDP & 8 & 71.9 & \bf 94.6 & 55.2 & \bf 82.0 & 35.1 & \bf 72.0 \\
Bulgarian-BTB & 8 & \bf 99.1 & 99.1 & 94.0 & \bf 95.2 & \bf 91.3 & 85.6 \\
Croatian-SET & 8 & 98.1 & \bf 98.3 & 91.7 & \bf 93.0 & \bf 87.4 & 84.4 \\
Danish-DDT & 8 & \bf 98.0 & 97.9 & 88.4 & \bf 89.3 & \bf 86.4 & 81.7 \\
Estonian-EDT & 8 & \bf 97.4 & 97.4 & 88.1 & \bf 88.9 & \bf 85.4 & 79.7 \\
Galician-CTG & 8 & \bf 97.7 & 97.7 & 85.4 & \bf 86.4 & \bf 83.0 & 81.1 \\
Galician-TreeGal & 8 & 94.2 & \bf 97.2 & 78.8 & \bf 86.2 & 73.8 & \bf 78.5 \\
Greek-GDT & 8 & \bf 98.0 & 97.9 & 91.8 & \bf 93.7 & \bf 89.7 & 87.6 \\
Slovak-SNK & 8 & 97.0 & \bf 97.5 & 91.4 & \bf 94.7 & \bf 88.8 & 84.4 \\
Slovenian-SSJ & 8 & \bf 98.7 & 98.6 & 93.2 & \bf 95.2 & \bf 91.7 & 88.3 \\
Slovenian-SST & 8 & \bf 95.0 & 94.8 & 66.3 & \bf 78.2 & 61.4 & \bf 63.3 \\
\rowcolor{lightgray} \textbf{AVERAGE} & & 95.0 & \bf 97.4 & 84.0 & \bf 89.3 & 79.5 & \bf 80.6 \\

Finnish-FTB & 9 & \bf 96.7 & 96.3 & 90.9 & \bf 91.5 & \bf 88.5 & 77.7 \\
Finnish-TDT & 9 & \bf 97.5 & 97.3 & 91.2 & \bf 91.3 & \bf 88.9 & 81.6 \\
Hebrew-HTB & 9 & \bf 97.5 & 97.2 & 82.5 & \bf 92.0 & 78.7 & \bf 85.2 \\
Indonesian-GSD & 9 & \bf 93.8 & 93.6 & 86.2 & \bf 86.5 & \bf 80.1 & 75.5 \\
Korean-GSD & 9 & \bf 96.5 & 96.4 & \bf 88.8 & 88.7 & \bf 85.3 & 78.0 \\
Korean-Kaist & 9 & \bf 95.7 & 95.6 & \bf 88.8 & 88.6 & \bf 86.9 & 80.0 \\
Norwegian-Bokmaal & 9 & 98.4 & \bf 98.5 & 93.0 & \bf 93.7 & \bf 91.4 & 85.7 \\
Persian-Seraji & 9 & \bf 97.9 & 97.9 & 91.3 & \bf 91.7 & \bf 88.4 & 84.8 \\
Romanian-RRT & 9 & 98.0 & \bf 98.0 & 91.5 & \bf 92.4 & \bf 87.2 & 83.4 \\
Serbian-SET & 9 & 97.5 & \bf 98.7 & 92.0 & \bf 94.7 & \bf 88.4 & 87.2 \\
Turkish-IMST & 9 & \bf 96.8 & 95.7 & 73.8 & \bf 74.5 & \bf 67.9 & 60.7 \\
Vietnamese-VTB & 9 & 89.5 & \bf 91.3 & 66.6 & \bf 73.9 & 59.1 & \bf 60.9 \\
\rowcolor{lightgray} \textbf{AVERAGE} & & 96.3 & \bf 96.4 & 86.4 & \bf 88.3 & \bf 82.6 & 78.4 \\

Arabic-PADT & 10 & 96.7 & \bf 96.8 & 82.6 & \bf 87.8 & 78.6 & \bf 80.5 \\
Catalan-AnCora & 10 & 98.9 & \bf 98.9 & 93.7 & \bf 94.3 & \bf 91.6 & 89.4 \\
Czech-CAC & 10 & 99.3 & \bf 99.3 & \bf 93.7 & 93.5 & \bf 91.6 & 85.9 \\
Czech-FicTree & 10 & 98.6 & \bf 98.6 & \bf 94.5 & 94.4 & \bf 92.0 & 84.5 \\
Czech-PDT & 10 & 99.1 & \bf 99.2 & 93.5 & \bf 94.0 & \bf 91.7 & 86.3 \\
Dutch-Alpino & 10 & 96.3 & \bf 97.3 & 92.1 & \bf 94.0 & \bf 89.7 & 86.5 \\
Dutch-LassySmall & 10 & \bf 96.6 & 96.5 & 89.6 & \bf 93.3 & \bf 87.0 & 82.9 \\
Hungarian-Szeged & 10 & \bf 96.6 & 96.6 & 87.4 & \bf 89.0 & \bf 82.8 & 80.4 \\
Swedish-LinES & 10 & \bf 97.4 & 97.2 & 87.5 & \bf 88.1 & \bf 84.1 & 78.2 \\
Swedish-Talbanken & 10 & 98.1 & \bf 98.2 & 91.1 & \bf 92.0 & \bf 88.8 & 83.9 \\
Ukrainian-IU & 10 & 97.7 & \bf 97.9 & 91.1 & \bf 91.3 & \bf 88.7 & 84.0 \\
\rowcolor{lightgray} \textbf{AVERAGE} & & 97.8 & \bf 97.9 & 90.6 & \bf 92.0 & \bf 87.9 & 83.9 \\

Chinese-GSD & 11 & 95.1 & \bf 96.5 & 83.3 & \bf 88.9 & 79.4 & \bf 81.5 \\
Italian-ISDT & 11 & 98.4 & \bf 98.7 & 94.0 & \bf 94.9 & \bf 92.3 & 88.5 \\
Italian-PoSTWITA & 11 & \bf 96.8 & 96.4 & 84.0 & \bf 86.1 & \bf 80.1 & 77.2 \\
Japanese-GSD & 11 & \bf 98.4 & 97.7 & 89.7 & \bf 94.1 & \bf 87.9 & 87.9 \\
Polish-LFG & 11 & 98.8 & \bf 98.8 & 96.6 & \bf 96.9 & \bf 95.0 & 82.5 \\
Polish-SZ & 11 & 98.3 & \bf 98.6 & 94.4 & \bf 94.9 & \bf 92.3 & 83.1 \\
Portuguese-Bosque & 11 & \bf 97.3 & 97.1 & 90.6 & \bf 93.0 & \bf 88.4 & 86.4 \\
\rowcolor{lightgray} \textbf{AVERAGE} & & 97.6 & \bf 97.7 & 90.4 & \bf 92.7 & \bf 87.9 & 83.9 \\

French-GSD & 12 & 97.6 & \bf 97.9 & 90.6 & \bf 92.3 & \bf 88.0 & 86.2 \\
French-Sequoia & 12 & 98.9 & \bf 99.1 & 92.0 & \bf 94.6 & \bf 90.5 & 89.0 \\
French-Spoken & 12 & 96.1 & \bf 96.7 & 80.4 & \bf 86.2 & \bf 75.8 & 74.7 \\
German-GSD & 12 & 94.9 & \bf 95.0 & 84.8 & \bf 87.3 & \bf 80.7 & 77.6 \\
Russian-SynTagRus & 12 & 99.0 & \bf 99.1 & 94.2 & \bf 94.7 & \bf 92.9 & 87.9 \\
Russian-Taiga & 12 & \bf 95.1 & 94.7 & 79.4 & \bf 80.8 & \bf 73.2 & 68.6 \\
Spanish-AnCora & 12 & 98.9 & \bf 99.0 & 92.8 & \bf 93.5 & \bf 91.0 & 88.5 \\
\rowcolor{lightgray} \textbf{AVERAGE} & & 97.2 & \bf 97.4 & 87.8 & \bf 89.9 & \bf 84.6 & 81.8 \\

English-EWT & 14 & 96.6 & \bf 96.7 & 87.6 & \bf 92.0 & \bf 85.4 & 81.9 \\
English-GUM & 14 & 96.3 & \bf 96.7 & 87.5 & \bf 91.8 & \bf 84.6 & 83.4 \\
English-LinES & 14 & 97.1 & \bf 97.4 & 86.1 & \bf 86.8 & \bf 82.0 & 77.6 \\
\rowcolor{lightgray} \textbf{AVERAGE} & & 96.7 & \bf 96.9 & 87.1 & \bf 90.2 & \bf 84.0 & 81.0 \\
\bottomrule
\end{tabular}
}
\caption{Breakdown of POS tagging and parsing experiment in \cref{sec:is-mbert-equal}. \label{tab:breakdown-ud}}
\end{center}
\end{table*}
}